\def\year{2023}\relax
\documentclass[letterpaper]{article} 
\usepackage{aaai22}  
\usepackage{times}  
\usepackage{helvet}  
\usepackage{courier}  
\usepackage[hyphens]{url}  
\usepackage{graphicx} 
\usepackage{booktabs}
\usepackage[dvipsnames]{xcolor}
\usepackage{amsmath,bm}
\usepackage{multirow}
\usepackage{amssymb}
\usepackage{float}
\usepackage{appendix}

\usepackage{cite} 
\usepackage[colorlinks=false,linkcolor=black,citecolor=black,urlcolor=blue,]{hyperref}

\usepackage{natbib}  
\usepackage{caption} 
\DeclareCaptionStyle{ruled}{labelfont=normalfont,labelsep=colon,strut=off} 
\usepackage{algorithm}
\usepackage{algorithmic}

\usepackage{newfloat}
\usepackage{listings}
\floatstyle{ruled}
\newfloat{listing}{tb}{lst}{}
\floatname{listing}{Listing}
\title{CAB: Empathetic Dialogue Generation with Cognition, Affection and Behavior}
\author{
    Pan Gao\textsuperscript{\rm 1},
    Donghong Han\textsuperscript{\rm 1,2}\thanks{Corresponding Author},
    Rui Zhou\textsuperscript{\rm 3},
    Xuejiao Zhang\textsuperscript{\rm 1},
    Zikun Wang\textsuperscript{\rm 1}\\
}

\affiliations{
    \textsuperscript{\rm 1}School of Computer Science and Engineering, Northeastern University, Shenyang, China\\
    \textsuperscript{\rm 2}Key Laboratory of Intelligent Computing in Medical Image of Ministry of Education, \\ Northeastern University, Shenyang, China\\
     \textsuperscript{\rm 3}Swinburne University of Technology, Australia\\
handonghong@cse.neu.edu.cn,  rzhou@swin.edu.au, \{1875917446, 2318821225, 2508954704\}@qq.com\\

}

\usepackage{bibentry}

\begin{document}
\maketitle

\begin{abstract}
Empathy is an important characteristic to be considered
when building a more intelligent and humanized dialogue agent. However, existing methods did not fully comprehend empathy as a complex
process involving three aspects: cognition, affection and behavior. In this
paper, we propose CAB, a novel framework that takes a comprehensive
perspective of cognition, affection and behavior to generate empathetic
responses. For cognition, we build paths between critical keywords in the
dialogue by leveraging external knowledge. This is because keywords in
a dialogue are the core of sentences. Building the logic relationship between keywords, which is overlooked by the majority of existing works,
can improve the understanding of keywords and contextual logic, thus
enhance the cognitive ability. For affection, we capture the emotional
dependencies with dual latent variables that contain both interlocutors’
emotions. The reason is that considering both interlocutors’ emotions
simultaneously helps to learn the emotional dependencies. For behavior,
we use appropriate dialogue acts to guide the dialogue generation to enhance
the empathy expression. Extensive experiments demonstrate that
our multi-perspective model outperforms the state-of-the-art models in both
automatic and manual evaluation.
\end{abstract}
\section{Introduction}
Empathy is the ability to understand others’ feelings, and respond appropriately to their situations \cite{hcr:2007MDEmpathy}. Previous studies have shown that empathetic dialogue
models can improve user’s satisfaction in several areas, such as customer service \cite{c:2021LZSupport}, healthcare community \cite{hcr:2021WWCASS} and etc. Therefore, how to successfully
implement empathy becomes one of the key issues to build an intelligent and
considerate agent. In recent years, many studies have been conducted on the \begin{figure}[t]
\centering
\includegraphics[width=1.0\columnwidth]{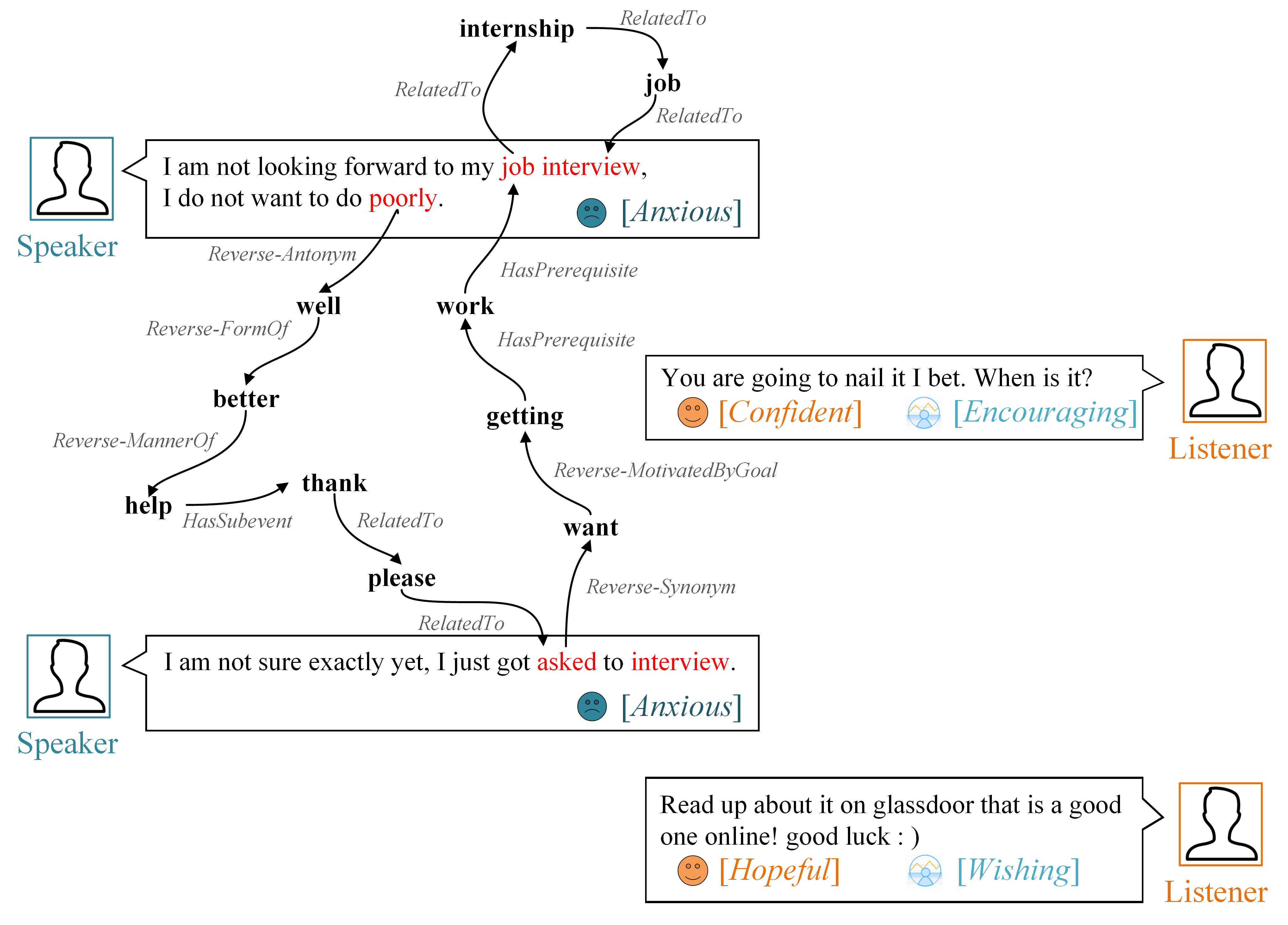} 
\caption{A dialogue from the \emph{EmpatheticDialogues} dataset. The cognitive ability is improved by retrieving entities (bold in black) and relationships (grey) from \emph{ConceptNet} and building paths between keywords (red) to generate a high quality response under the influence of \emph{anxious} and \emph{confident} emotions and \emph{wishing} dialogue act.}
\label{fig:fig1}
\end{figure}
\vspace{0.0cm}task of empathetic dialogue generation, which are mainly divided into two categories: One is to enhance the understanding of a user’s situation and emotion by
leveraging knowledge from one or more external knowledge bases~\cite{hcr:2021WLMultiCausality,c:2022SZCEM,hcr:2021LMGPT,c:2022LLBridging} or adding emotion causes as prior emotion knowledge~\cite{c:2021GLImCause,hcr:2021WLMultiCausality}. This is to improve
the cognitive ability. The issue of the existing work is that they overlook the
importance of paths between users’ critical keywords, which can actually reflect
the contextual logic in the conversation. Although some studies \cite{hcr:2021WLMultiCausality} build paths between emotion concepts and cause concepts, they mainly focus on the causality aspect and ignore the fact that paths between any keywords can help. The second category is to design emotion strategies, such as mixture of experts \cite{c:2019LMMoEL}, emotion mimicry \cite{c:2020MHMIME} and multi-resolution emotions \cite{c:2020LCEmpDG} to generate appropriate responses from the affection aspect. Unfortunately, these studies learn to respond properly mainly according to the speaker’s emotion rather than both
interlocutors’ emotions. In this paper, we aim to improve the aforementioned
weak aspects of the existing works to help advance the study of empathetic
dialogue generation.

Psychological research shows that empathy is a complex mental process involving three aspects of interlocutors: cognition, affection and behavior \cite{hcr:2009LWreview}.
Specifically, cognitive empathy refers to the ability to understand and interpret
a user’s situation \cite{hcr:2018EBTherapist}; affective empathy is an emotional reaction based on differentiating the emotions of oneself and others \cite{hcr:2009LWreview}; behavioral empathy means verbal or non-verbal forms of communication used in the empathetic dialogue \cite{hcr:1977GEmpathy}.
Among the existing works, some only consider the aspects of congition and affection \cite{c:2022SZCEM,c:2021ZLCOMAE}; others mainly consider the aspect of behavior \cite{c:2020WPTaxonomy,c:2022EmpHi}. None of the existing works had comprehensively considered all the three aspects (cognition,
affection, behavior), which we believe are all important. In the following, we
elaborate in detail with the example in Figure~\ref{fig:fig1}. The dialogue in Figure~\ref{fig:fig1} shows
that 1) \textbf{Cognition}: The speaker is \emph{anxious}  about attending a job interview. In the first turn, 
there exists a path between \textless \emph{job}, \emph{interview}\textgreater\ with internship as a bridge to enhance the understanding of the keywords and the context. In the next turn, the paths between \textless \emph{poorly}, \emph{asked}\textgreater\ and \textless \emph{asked}, \emph{job}\textgreater\ are built to alleviate the problem that it is difficult to capture the contextual logic based on limited context. Thus, it can be seen that the paths, which establish the relationships
between utterances, are critical to improve the cognitive ability. 2) \textbf{Affection}: In interpersonal conversations, responses are usually influenced by both interlocutors' emotions \cite{c:2019GMEmotionRecognition}. As shown in Figure~\ref{fig:fig1}, in the second turn, instead of both sides falling into anxiety, the listener is able to perceive the speaker's emotion 
and accept the emotion difference between them, 
thus generating a response with more positive emotion (\emph{hopeful}). Therefore, how to learn the emotional dependencies between the context and target response based on both participants’ emotions 
is critical for responding properly. 3) \textbf{Behavior}: Appropriate dialogue acts are used as communicative form to enhance empathy expression. For example, the listener inspires the speaker by \emph{encouraging} and make the speaker relaxed by \emph{wishing}. Different from Welivita and Pu \citeyearpar{c:2020WPTaxonomy}, we
choose dialogue act of 8 categories (\emph{agreeing}, \emph{acknowledging}, \emph{encouraging}, \emph{consoling}, \emph{sympathizing}, \emph{suggesting}, \emph{questioning}, \emph{wishing}). In this way, we can guide
dialogue generation better.

To this end, we propose a novel Empathetic Dialogue Generation model including aspects of \textbf{C}ognition, \textbf{A}ffection and \textbf{B}ehavior (\textbf{CAB}) to achieve a comprehensive empathetic dialogue task. Specifically, since keywords are important to understand the contextual logic, our model builds paths between keywords through multi-hop commonsense reasoning to enhance the cognitive ability. Conditional variational auto-encoder (CVAE) model 
with dual latent variables is built based on both interlocutors' emotions, and then the dual latent variables are injected into the decoder together with the dialogue act features to produce empathetic responses from the perspective of affection and behavior.
Our contributions are summarized as follows:
\begin{itemize}
\item To the best of our knowledge, we are the first to propose a novel framework for empathetic dialogue generation based on psychological theory from three perspectives: cognition, affection and behavior.
\item We propose a context-based multi-hop reasoning method, in which paths are established between keywords to acquire implicit knowledge and learn contextual logic.
\item We present a novel CVAE model, which introduces dual latent variables to learn the emotional dependencies between the context and target responses. After that, we incorporate the dialogue act features into the decoder to guide the generation.
\item Experiments demonstrate that 
CAB generates more relevant and empathetic responses compared with the state-of-the-art methods.$\footnote{Code and data are available at https://github.com/geri-emp/CAB}$

\end{itemize}

\section{Related Work}
\subsection{Empathetic Dialogue Generation}
 Recently, there has been numerous works in the task of empathetic dialogue generation proposed by Rashkin et al. \citeyearpar{c:2019RSBenchmark}. Lin et al. \citeyearpar{c:2019LMMoEL} assign different decoders for various emotions, and fuse the output of each decoder with users' emotion weights. Majumder et al. \citeyearpar{c:2020MHMIME} adopt emotion stochastic sampling and emotion mimicry to respond to positive or negative emotions for generating empathetic responses. Li et al. \citeyearpar{c:2020LCEmpDG} construct an interactive adversarial learning network considering multi-resolution emotions and user feedback. Liu et al. \citeyearpar{c:2021LDResponses} incorporate anticipated emotions into response generation via reinforcement learning. Gao et al.  \citeyearpar{c:2021GLImCause} introduce emotion cause to better understand the user's emotion. However, all of the above methods only consider the user's emotion and ignore the influence between both interlocutors’ emotions in the dialogue.

Several studies have incorporated external knowledge into empathetic dialogue generation. Li et al. \citeyearpar{c:2022LLBridging} employ multi-type knowledge to explore implicit information and construct an emotional context graph to improve emotional perception. Liu et al. \citeyearpar{hcr:2021LMGPT} prepend the retrieved knowledge triples to the gold responses in order to get proper responses. However, these approaches retrieve knowledge triples without fully considering the contextual meaning of the words. Although Wang et al. \citeyearpar{hcr:2021WLMultiCausality} adopt \emph{ConceptNet} to explore the emotional causality by commonsense reasoning between the emotion clause and the cause clause, the logical relationships between other utterances may be ignored. Sabour et al. \citeyearpar{c:2022SZCEM} use \emph{ATOMIC} for commonsense reasoning to better understand the user's situation and feeling, but reasoning on a whole dialogue history may neglect the important role of keywords in the context. To overcome the above proposed shortcomings, we propose a context-based multi-hop commonsense reasoning method to enrich contextual information and reason about the logical relationships between utterances.

\begin{figure*}[t]
\centering
\includegraphics[width=1.8\columnwidth]{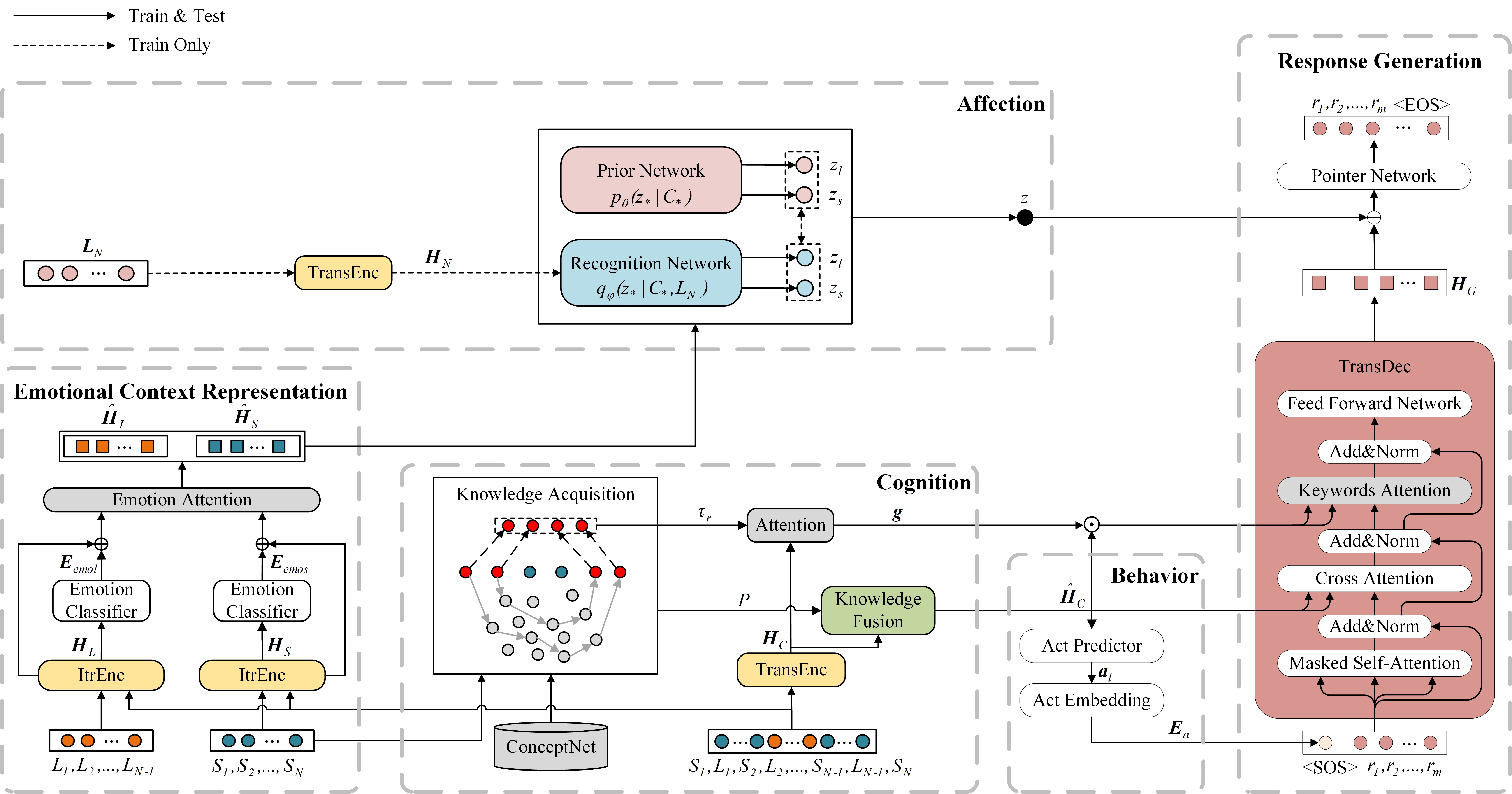}
\caption{The overall architecture of CAB.}
\label{fig:model}
\end{figure*}
\vspace{0.0cm}

\subsection{CVAE for Dialogue Generation}
The Seq2Seq framework is widely used since neural network models are becoming mainstream in NLP. However, this method tends to generate generic responses. CVAE \cite{c:2015SLCVAE} uses deep neural networks to fit probability distributions, and with its advantages in modelling text diversity, it can tackle the issue of generating dull responses to a certain extent. Ruan et al. \citeyearpar{hcr:2021RLAutoencoder} regularize the latent space of CVAE by emotion labels to incorporate emotion information into latent variables. Zhao et al. \citeyearpar{c:2017ZZDiscourse-level} use CVAE to learn the distribution of dialogue intent to capture discourse-level diversity. Lin et al. \citeyearpar{hcr:2020LWDiverse} construct fine-grained latent variables to enhance the Transformer decoder. Du et al. \citeyearpar{c:2018DLGeneration} incorporate a series of latent variables into the autoregressive decoder to model the multi-modal distribution of text sequences. These methods cannot be directly applied to our work as they are unable to capture both interlocutors' emotions. Therefore, we introduce dual latent variables in Transformer-based CVAE to learn emotional dependencies in empathetic dialogue generation.

\section{Method}
\subsection{Task Formulation and Overview}
In empathetic dialogue generation, each dialogue consists of a dialogue history $C=[S_1,L_1,S_2,L_2,…,S_{N-1},L_{N-1},S_{N}]$ of 2$N$-1 utterances and a gold empathetic response $L_N=[w_N^1,w_N^2,…,w_N^n]$ of $n$ words, where $S_i$ and $L_i$ denote the $i$-th utterance of speaker and listener respectively. Our goal is to generate a fluent, appropriate and empathetic response $R=[r_1,r_2,…,r_m]$ based on the dialogue history $C$, the speaker's emotion $e_s$, the listener's emotion $e_l$, and the listener's dialogue act $a_l$. 

We provide a overview of CAB in Figure~\ref{fig:model}, which consists of five components: \textbf{a) Emotional Context Representation}. The predicted emotions, $e_s$ and $e_l$, are fed into context $C$ by emotional context encoder to obtain the emotional context representation $\bm{\hat{H}}_S$ and $\bm{\hat{H}}_L$; \textbf{b) Affection}. Then prior network and posterior network capture dual latent variables $\bm{z}_s$ and $\bm{z}_l$, based on $\bm{\hat{H}}_S$ and $\bm{\hat{H}}_L$ in the test and training phase; \textbf{c) Cognition}. To build paths $P$, we leverage \emph{ConceptNet} to acquire external knowledge and incorporate it into $C$ to obtain a knowledge-enhanced context representation $\bm{\hat{H}}_C$; \textbf{d) Behavior}. The dialogue act features $\bm{E}_a$ are distilled based on a predictor and the embedding layer; \textbf{e) Response Generation}. 
The three-stage decoder  generates an 
empathetic response $R$ based on the aspects of affection, cognition and behavior.

We evaluate the model on \emph{EmpatheticDialogues} \cite{c:2019RSBenchmark}, which is a publicly available benchmark dataset for empathetic dialogue generation. However, dialogues in this dataset do not contain emotion label and dialogue act label for each listener's utterance, and thus we annotate emotion and dialogue act by Emoberta \cite{hcr:2021KVEmoBERTa} and EmoBERT \cite{c:2020WPTaxonomy} (See in Appendix A),
respectively, to support the studies in this paper.

\subsection{Emotional Context Encoder}
\subsubsection{Input Representation.} We divide the dialogue history into two segments $C_S=[S_1,S_2,…,S_N]$ and $C_L=[L_1,L_2,…,L_{N-1}]$. Following the previous work \cite{c:2019LMMoEL}, we concatenate the utterances in $C_S$, $C_L$, $C$  and prepend a special token [CLS] to gain the speaker context, listener context and global context respectively. The three sequences and the gold response $L_N$ are passed through the embedding layer to obtain $\bm{E}_S$, $\bm{E}_L$, $\bm{E}_C$ and $\bm{E}_N$. We feed $\bm{E}_C$ and $\bm{E}_N$ into the Transformer encoder (\textbf{TransEnc}) to obtain the context representation $\bm{H}_C$ and the response representation $\bm{H}_N$. Meanwhile, $\bm{E}_S$ and $\bm{E}_L$ as query, and $\bm{E}_C$ as key and value are fed into the Transformer-based inter-encoder (\textbf{ItrEnc}) to derive the speaker and listener contextual utterances representation $\bm{H}_S$ and $\bm{H}_L$:
\begin{equation}
\bm{H}_C={\rm TransEnc}
(\bm{E}_C), \bm{H}_N={\rm TransEnc}(\bm{E}_N) 
\end{equation}
\begin{equation}
\begin{split}
\bm{H}_S={\rm ItrEnc}(\bm{E}_S,\bm{E}_C), 
\bm{H}_L={\rm ItrEnc}(\bm{E}_L,\bm{E}_C)
\end{split}
\end{equation}
where $\rm TransEnc$ and $\rm ItrEnc$ are encoder and inter-encoder of output size $d$ based on Transformer, and $\bm{H}_C \in \mathbb{R}^{c\times d}$, $\bm{H}_N  \in \mathbb{R}^{n\times d}$, $\bm{H}_S  \in \mathbb{R}^{s\times d}$, $\bm{H}_L  \in \mathbb{R}^{l\times d}$, with $c$, $n$, $s$ and $l$ being the length of the above four sequences.
\subsubsection{Emotion Classification.}To understand the emotions of the speaker and the listener, we project the hidden representations of the token [CLS] from $\bm{H}_S$ and $\bm{H}_L$, namely $\bm{H}_{S0}$ and $\bm{H}_{L0}$, into the emotion category distribution $P_{s}$ and $P_{l}$:
\begin{equation}
\begin{aligned}
P_{s}(\bm{e}_s|C_S)&= {\rm Softmax}(\bm{W}_s \bm{H}_{S0}),\\
P_{l}(\bm{e}_l|C_L)&= {\rm Softmax}(\bm{W}_l \bm{H}_{L0}) 
\end{aligned}
\end{equation}
where $\bm{W}_s$, $\bm{W}_l \in \mathbb{R}^{k\times d}$ are learnable parameters, and $k$ is the number of emotion categories. During training, We optimize the model by minimizing the cross-entropy loss:
\begin{equation}
\begin{split}
\mathcal{L}_{s}={\rm -log}(P_{s} (\bm{e}_s^*)),
\mathcal{L}_{l}={\rm -log}(P_{l} (\bm{e}_l^*))
\end{split}
\end{equation}
where $\bm{e}_s^*$ and $\bm{e}_l^*$ denote the ground truth emotion label of the speaker and the listener respectively. Then we send $\bm{e}_s$ and $\bm{e}_l$ to the trainable emotion embedding layer to obtain the emotion states embedding matrix $\bm{E}_{emos}$, $\bm{E}_{emol} \in \mathbb{R}^{k\times de}$, and $de$ denotes the dimension of emotion embedding.
\subsubsection{Emotion Self-Attention.} To make the latent variables in the next section incorporate both interlocutors' emotions, $\bm{H}_S$ and $\bm{H}_L$ are concatenated with $\bm{E}_{emos}$ and $\bm{E}_{emol}$  and fed into a self-attention layer followed by a linear layer to obtain the emotional context representation $\bm{\hat{H}}_S\in \mathbb{R}^{s\times d}$ and $\bm{\hat{H}}_L\in \mathbb{R}^{l\times d}$.
\subsection{Prior Network and Recognition Network (Affection)}
We introduce dual latent variables $\bm{z}_*\in\{\bm{z}_s,\bm{z}_l\}$ in CVAE, mapping the input sequences $C_*\in \{C_S,C_L\}$ into the output sequence $L_N$ via $\bm{z}_*$. In this section, taking speaker as an example, we illustrate how to realize the prior network and the recognition network.

The \textbf{prior network} $p_\theta (\bm{z}_s \vert C_S)$ is parameterized by 3-layer MLPs to compute the mean $\mu_s^\prime$ and variance $\sigma_s^{\prime2}$ of $\bm{z}_s\in\mathbb{R}^{dz}$. 
The network structure of the \textbf{recognition network} $q_\varphi (\bm{z}_s \vert C_S,L_N)$ is the same as that of the prior network, except that the input also includes $\bm{H}_N$. We sample $\bm{z}_s$ by reparameterization trick \cite{c:2014KWAuto-Encoding}.
\begin{equation}
\begin{split}
[\mu_s^\prime,\sigma_s^{\prime2}]= {\rm f}_{prior}^s (\bm{\hat{H}}_S),
[\mu_s,\sigma_s^2]= {\rm\bm f}_{recog}^s (\bm{\hat{H}}_S,\bm{H}_N)
\end{split}
\end{equation}
where ${\rm f_{prior}^s}$ and ${\rm f_{recog}^s}$ are speaker's prior network and recognition network; $dz$ is the dimension of the latent variable. In order to help the model learn the emotional dependencies based on both interlocutors' emotions, we fuse $\bm{z}_s$ and $\bm{z}_l$ due to the emotional similarity coefficient $\beta$ between $\bm{E}_{emos}$ and $\bm{E}_{emol}$ to obtain 
$\bm{z}=\beta\cdot \bm{z}_s+(1-\beta)\cdot \bm{z}_l$. 

The architecture of the listener's prior network and recognition network is the same as that of the speaker. CVAE model is trained by minimizing:
\begin{equation}
\begin{split}
\mathcal{L}(C_*,L_N;\theta,\varphi)=\alpha {\rm KL}(q_\varphi (\bm{z}_s\vert C_S,L_N)\Vert p_\theta(\bm{z}_s \vert C_S))\\
-\bm{E}_{q_\varphi (\bm{z}_s\vert C_S,L_N } [\log p_\theta(L_N \vert \bm{z}_s,C_S )] \\
+\alpha {\rm KL}(q_\varphi(\bm{z}_l\vert C_L,L_N)\Vert p_\theta(\bm{z}_l \vert C_L))\\
-\bm{E}_{q_\varphi (\bm{z}_l\vert C_L,L_N)} [\log p_\theta(L_N\vert \bm{z}_l,C_L)]
\end{split}
\end{equation}
where $\alpha\in[0,1]$ is the modified
\ coefficient for the ${\rm KL}$ term in the KL annealing strategy \cite{c:2016BVContinuous}.
\subsection{Knowledge Acquisition and Fusion (Cognition)}
\subsubsection{Knowledge Acquisition.}\label{subsec:knowledgeAcquisition} We first use the TextRank algorithm \cite{c:2004MTTextRank} to obtain 3-10 important keywords based on the length of the speaker's utterances $C_S$, and then 
filter $\bm {cw}$ words with noun, verb, adverb and adjective lexical properties among these keywords to get $\tau_{all}=\{h_1,h_2,…,h_{cw}\}$, more details in the Appendix B. Then we build paths as follows:

\textbf{Step a}. Take one keyword in $\tau_{all}$ as the origin keyword $h_i\in\tau_{all}$ and get $\bm{E}_i$ by feeding $h_i$ into the word embedding layer, then send $\bm{E}_i$ and $\bm{E}_s$ into equations~\ref{eq:7} to get $\bm{\delta}_i$, which is the contextual semantic features of $h_i$. The Top-$K$ knowledge triples in \emph{ConceptNet} $\omega_K= \{h_i,r_i^q,t_i^q,s_i^q\},q\in\{1,2,…,K\}$ associated with $h_i$ are retrieved based on score and removed relation set (See in Appendix B), where score is the sum of the cosine similarity score between $t_i^q$ and $\delta_i$ and confidence score $s_i^q$ that is scaled to a number in the interval [0,1] by minimum-maximum normalization.
\begin{equation}
\label{eq:7}
\begin{split}
\bm{\delta}_i={\rm ItrEnc}(\bm{E}_i,\bm{E}_s), i=1,2,…,cw
\end{split}
\end{equation}

\textbf{Step b}. To ensure that the triples are logically related to the other keywords $\tau_{other}=\{h_1,h_2,…,h_{i-1},h_{i+1},…,h_{cw}\}$, we first obtain the contextual semantic features $\bm{\delta}_j$ of $h_j\in\tau_{other}$ by $\rm ItrEnc$ like step a. After ranking the triples by cosine similarity score between $t_i^q$ and $\delta_j$, we filter $\omega_K$ to get Top-$k$ triples related to $h_j$, $\omega_k=\{h_i,r_i^p,t_i^p,s_i^p\}$, $p\in\{1,2,...,k\}$, $\omega_k\subseteq\omega_K$, and determine whether $t_i^p$ is the same as $h_j$. If the words are same, there exists a one-hop path $<h_i-r_i^p\rightarrow h_j (t_i^p)> $ between $h_i$ and $h_j$, and we add $h_i$ and $h_j$ to the final retained keywords set $\tau_r$ (e.g. red circles in Figure~\ref{fig:model}). Then the attention weight vector $\bm{g}\in \mathbb{R}^c$ is calculated for each word in $C$ with $\tau_r$ by the attention layer. Then $t_i^p$ is added to $\tau_{all}$ to continue finding the paths. Step a and b are repeated until all paths are found (the hops of paths do not exceed $M$) and $Num$ paths $P$ are retained (e.g. the paths connected by grey arrows in Figure~\ref{fig:model}).

\textbf{Step c}. The dialogues for which no path is found have special process. We first change the values of $K$, $k$ and $Num$, and repeat steps a and b. Finally, the remaining 4.8\% of the data for which the path is still not found is supplemented by finding a one-hop knowledge triple according to Step a. 
\subsubsection{Knowledge Fusion.}
We first convert the paths into sequences (See in Appendix B), e.g. \textless \emph{home} - \emph{RelatedTo} $\rightarrow$ \emph{heart} - \emph{UsedFor} $\rightarrow$ \emph{love}\textgreater\ into 'Home is related to heart. Heart is used for love.'. Then the sequences are fed into the two-layer Bi-GRU to obtain the knowledge representation $\bm{H}_k$. Finally, following previous work \cite{c:2022SZCEM}, we concatenate $\bm{H}_k$ with context at token-level to obtain the knowledge-enhanced context representation $\bm{\hat{H}}_C$:
\begin{equation}
\begin{split}
\bm{\hat{H}}_C^i= \bm{H}_C^i\oplus\alpha_i\bm{H}_k
\end{split}
\end{equation}
\begin{equation}
\begin{split}
\alpha_i={\rm Softmax}(\bm{H}_C^i\cdot(\bm{H}_k \bm{W}_k )^T)
\end{split}
\end{equation}
where $\oplus$ is the concatenation operation; $\bm{H}_k\in \mathbb{R}^{p\times2d}$; $p$ is the length of the knowledge sequence; $\bm{W}_k\in \mathbb{R}^{2d\times d}$ is the weight matrix of linear layer.
\subsection{Dialogue Act Predictor and Representation (Behavior)}
Dialogue acts as a form of communication can not only demonstrate the understanding of the speaker's situation, furthermore, it can express empathy by \emph{questioning}, \emph{sympathizing} and \emph{encouraging}, etc. The model uses the hidden state of token [CLS] from $\bm{\hat{H}}_C$ to predict dialogue act $\bm{a}_l$ to guide the response, and takes maximizing the probability of the
dialogue act with a cross-entropy loss function as objective. Then, $\bm{a}_l$ is fed into the embedding layer to learn the dialogue act embedding representation $\bm{E}_a\in \mathbb{R}^{r\times da}$. 
\begin{equation}
\begin{split}
P_{a}(\bm{a}_l |C,P)= {\rm Softmax}(\bm{W}_a \bm{\hat{H}}_{C0})
\end{split}
\end{equation}
\begin{equation}
\begin{split}
\mathcal{L}_{act}=-\log(P_{a} (\bm{a}_l^*))
\end{split}
\end{equation}
where $da$ and $r$ are the dimension of dialogue act embedding and the number of dialogue act categories respectively; $W_a\in \mathbb{R}^{r\times d}$ is learnable parameter; $\hat{H}_{C0}$ is the hidden state of [CLS] token; 
 $a_l^*$ is the truth dialogue act label.
\subsection{Response Generation}
Finally, the aforementioned information $\bm{E}_a$, $\bm{g}$, $\bm{z}$ and $\bm{\hat{H}}_{C}$ are used at the Transformer-based decoder (\textbf{TransDec}) through the following three stages. Here we adopt pointer network \cite{c:2015VFNetworks} to generate a empathetic response $R=[r_1,r_2,…,r_m]$.
\subsubsection{Incorporating dialogue act features.} We concatenate the dialogue act embedding $\bm{E}_a$  and the embedding of the start-of-sequence 
token  $\bm{E}_{SOS}$, then we use a linear transformation to obtain a new embedding:
\begin{equation}
\begin{split}
\bm{E}_{SOS}^\prime=\bm{W}_t(\bm{E}_a\oplus \bm{E}_{SOS})
\end{split}
\end{equation}
where $\bm{W}_t\in \mathbb{R}^{d\times (d+da)}$.
\subsubsection{Knowledge-enhanced decoder.} Since the keywords between the paths are essential for generation, we design a multi-head keywords attention, which takes the output of the cross-attention layer as query, the dot-product over $\bm{g}$ and $\bm{\hat{H}}_C$ as key and value. Then TransDec outputs the hidden state $\bm{H}_G$:
\begin{equation}
\begin{split}
\bm{H}_G={\rm TransDec}(\bm{g},\bm{\hat{H}}_C,\bm{E}_{SOS}^\prime\oplus \bm{E}_{r_{1:t-1}})
\end{split}
\end{equation}
where $\rm TransDec$ is the decoder based on Transformer, and $\bm{E}_{r_{1:t-1}}$ denotes the embeddings of the generated tokens.
\subsubsection{Incorporating latent variable.} In order to learn the emotional dependencies in the decoder, we concatenate $\bm{z}$ and $\bm{H}_G$ at token-level and use pointer network to output the probability distribution of each word in the vocabulary:
\begin{equation}
\begin{split}
P(r_t\vert r_{0:t-1},\bm{z},\bm{g},\bm{\Hat{H}}_C)={\rm PointGen}(\bm{H}_G^i \oplus \bm{z})
\end{split}
\end{equation}
where $\rm PointGen$ is the pointer network.
\subsection{Training Objectives}
To avoid vanishing
latent variable problem in CVAE, we use bag-of-word loss \cite{c:2017ZZDiscourse-level} in addition to the KL annealing strategy. The bag-of-word loss is:
\begin{equation}
\begin{split}
\mathcal{L}_{bow}=-E_{q_\varphi (\bm{z}\vert C,L_N ) }\log p(L_{bow}\vert \bm{z},C) ] 
\end{split}
\end{equation}
where $L_{bow}$ is the bag-of-words representation of $L_N$.
The total loss of CAB is on the weighted sum of the five mentioned losses:
\begin{equation}
\begin{split}
\mathcal{L}=\gamma_1 \mathcal{L}_{s}+\gamma_2 \mathcal{L}_{l}+\gamma_3 \mathcal{L}_{a}+
\gamma_4 \mathcal{L}(C_*,C_N;\theta,\varphi)+\gamma_5 \mathcal{L}_{bow} 
\end{split}
\end{equation}
where $\gamma_1 = 0.03$, $\gamma_2 = 0.1$,  $\gamma_3 = 0.05$, $\gamma_4 = 1$ and $\gamma_5 = 0.1$ are hyper-parameters that we use to control the impact of the five losses. Specifically, the lower weights coefficient before the $\gamma_1$, $\gamma_2$ and $\gamma_3$ are to optimize all losses at the same time, which does not mean that these losses have contributed less.

\section{Experiments}
\subsection{Baselines}
We select the following state-of-the-art models for comparison:
1) \textbf{Transformer} \cite{c:2017VSAttention}: The vanilla Transformer with the pointer network, which is trained to optimize the negative log-likelihood loss. 2) \textbf{Multi-Transformer(Multi-Trans)} \cite{c:2019RSBenchmark}: A variant of Transformer that includes emotion classification loss in addition to the generation loss to jointly optimize the model. 3) \textbf{MOEL} \cite{c:2019LMMoEL}: A model that includes several Transformer decoders, each of which learns how to respond to an emotion, and the outputs are softly combined to generate responses. 4) \textbf{MIME} \cite{c:2020MHMIME}: A model adopting emotion mimicry and emotion clusters to deal with positive or negative emotions. 5) \textbf{EmpDG} \cite{c:2020LCEmpDG}: A generative adversarial network that considers multi-resolution emotion and introduces discriminators to supervise the training in semantics and emotion. 6) \textbf{KEMP} \cite{c:2022LLBridging}: A model that uses multi-type knowledge to help understand and express emotions, and learns emotional dependencies with emotional cross-attention mechanisms. 7) \textbf{CEM} \cite{c:2022SZCEM}: A method for generating empathetic responses by leveraging commonsense to improve the understanding of interlocutors’ situations and feelings.

In order to verify the effectiveness of each component, the following ablation experiments are conducted: 1) \textbf{w/o Cog}: The knowledge acquisition and fusion are removed, and the multi-head keywords attention mechanism is removed from TransDec. 2) \textbf{w/o Aff}: The CAB model without considering the listener's emotion. Thus the whole dialogue history without adding paths is fed into speaker's CVAE and emotion classifier. 3) \textbf{w/o Beh}: The classification of dialogue act and the dialogue act features fused at the decoder are removed.

\subsection{Implementation Details}
We implement all models in PyTorch \citeyearpar{c:2019PGPyTorch} with GeForce GTX 3090 GPU, and train models using Adam optimization \cite{c:2015KBOptimization} with a mini-batch size of 16. All common hyper-parameters are the same as the work in Lin et al \citeyearpar{c:2019LMMoEL}. We adopt 300-dimensional pre-trained GloVE vectors \cite{c:2014PSGlove} to initialize the word embeddings, which are shared between the encoders and the decoder. The hidden size is 300 everywhere, and the size of latent variable is 200. We use the KL annealing of 15,000 batches to achieve the best performance. During test, the batch size is 1 and the maximum greedy decoding steps is 50. The size of train/valid/test is 40244/5738/5259.

\subsection{Evaluation Metrics}
\subsubsection{Automatic Evaluation.} We choose PPL \cite{hcr:2015VLConversational}, Distinct-1, Distinct-2 
\cite{c:2016LGDiversity-Promoting} as our main automatic metrics. \textbf{PPL} is used to estimate the probability of a sentence based on each token and can measure the quality of a model in general. \textbf{Distinct-1} and \textbf{Distinct-2} measure the proportion of unique n-grams in the generated responses, and are used to measure the diversity of responses. Since emotion accuracy of speaker/listener (\textbf{EmoSA}/\textbf{EmoLA}) reflects the understanding of both interlocutors' emotions and dialogue act accuracy (\textbf{ActA}) can determine whether the appropriate dialogue acts are chosen to produce responses, we also report these three metrics. 

\vspace{0.0cm}
\begin{table}[t]
\centering
\resizebox{1.0\columnwidth}{!}{
\begin{tabular}{ccccccc}
\toprule
\multirow{1}{*}{\textbf{Models}} & \textbf{PPL} & \textbf{DIST-1} & \textbf{DIST-2} & \textbf{EmoSA} & \textbf{EmoLA} & \textbf{ActA} \\
\midrule
Transformer & 34.11 & 0.49 & 1.91 & - & - & - \\
Multi-Trans & 36.42 & 0.43 & 1.85 & 28.91 & - & - \\
MOEL & 36.59 & 0.60 & 3.12 & 32.33 & - & - \\
MIME & 37.52 & 0.32 & 1.22 & 34.88 & - & - \\
EmpDG & 37.37 & 0.45 & 1.89 & 32.45 & - & - \\
KEMP & 36.39 & 0.66 & 3.08 & 36.57 & - & -  \\
CEM & 36.11 & 0.66 & 2.99 & 39.07 & - & - \\
\midrule
\textbf{CAB} & 34.36 & \textbf{1.13} & \textbf{4.23} & \textbf{40.52} & \textbf{72.23} & 41.72 \\
w/o Cog & \textbf{33.88} & 0.94 & 3.33 & 39.42 & 71.82 & \textbf{43.09} \\
w/o Aff & 34.98 & 1.12 & 3.97 & 34.25 & - & 37.25 \\
w/o Beh & 34.79 & 1.06 & 3.83 & 40.05 & 72.20 & -\\
\bottomrule
\end{tabular}}
\caption{Results of the automatic evaluation, and w/o Cog/Aff/Beh indicate ablation experiments and the best results of all models are bold.}
\label{tab:table1}
\vspace{-0.3cm}
\end{table}
\vspace{0.0cm}

\subsubsection{Human Evaluation.} The human evaluation includes two parts: human ratings and A/B test, which are completed by three annotators from a third-party company. For the first part, we randomly select 100 dialogues from our model as well as the baseline models, and then ask the annotators to score the generated responses of each model on 1 to 5 point scale in terms of empathy, relevance and fluency. \textbf{Empathy} is the ability to understand the user's emotion and experience. \textbf{Relevance} evaluates whether the responses are relevant to the dialogue history. \textbf{Fluency} measures whether the generated response is grammatically correct and readable. For the second part, we re-sample dialogues to obtain 100 context-response pairs for CAB vs. \{MoEL, MIME, EmpDG, KEMP, CEM\}. The annotators choose the better one or a Tie if they think both responses are good. To ensure fairness, each group of A/B test uses a distinct dialogue context.

\subsection{Results and Analysis}
\subsubsection{Automatic Evaluation Results.} The overall automatic evaluation results are shown in the Table~\ref{tab:table1}. Our model CAB outperforms the baselines on all metrics significantly. The lower ppl score implies that CAB has a higher quality of generation generally, reflecting the importance of considering empathy from multi-perspective. The remarkable improvements in distinct-1 and distinct-2 suggest that the introduction of external knowledge can be beneficial in improving the understanding of dialogue history and thus generating a wider variety of response. The higher accuracy of emotion classification verifies the validity of modelling both interlocutors' emotions separately.

As in the bottom part of Table~\ref{tab:table1}, we also conduct ablation experiments to explore the effect of each component. When commonsense knowledge is removed \textbf{(w/o Cog)}, all metrics decrease except for PPL, especially Distinct-1 and Distinct-2, suggesting that the paths capture additional information to enhance cognitive ability, thus improving the quality and diversity of responses. The increasing PPL score may be due to the introduction of knowledge, which may have an impact on the fluency of the generated responses. In addition, we find that only considering the speaker's emotion \textbf{(w/o Aff)} yields lower emotion accuracy and higher ppl score, and thus it is difficult to generate appropriate responses without understanding both interlocutors' emotions exactly. All metrics decrease when the generated responses lack the guiding of the dialogue acts \textbf{(w/o Beh)}, indicating the emphasis of the dialogue acts in improving empathy.

\vspace{0.0cm}
\begin{table}[t]

\centering
\begin{tabular}{ccccc}
\toprule
\textbf{Models} &\textbf{Empathy} & 
\textbf{Relevancy} & \textbf{Fluency} & \textbf{$\kappa$}\\
\midrule
MOEL & 3.29 & 2.94 & 4.24 & 0.24\\
MIME & 3.43 & 3.18 & 4.34 & 0.22\\
EmpDG & 3.41 & 3.04 & 4.29 & 0.27\\
KEMP & 3.26 & 3.02 & 4.34 & 0.26 \\
CEM & 3.44 & 3.10 & \textbf{4.40} & 0.26\\
\textbf{CAB} & \textbf{3.52} & \textbf{3.30} & 4.30 & 0.20\\
\bottomrule
\end{tabular}
\caption{Results of the human ratings. The Fleiss-Kappa $\kappa$ of each model indicates that the multiple annotators almost reach a fair agreement.}
\label{tab:table2}
\end{table}

\subsubsection{Human Evaluation Results.} Table~\ref{tab:table2} illustrates that CAB yields the highest scores in empathy and relevance. It proves that incorporating knowledge enriches the context and captures more vital information, which makes the responses more relevant to the context. Considering the common influence of both interlocutors' emotions can learn the emotional dependencies between context and responses better, resulting in more empathetic responses. The dialogue act can guide the model to generate appropriate empathetic responses. Although CAB is not as good as the other models in fluency, the difference is not significant. Additionally, the human A/B test results from Table~\ref{tab:table3} show that the annotators consider our responses to be better and more acceptable.

\subsubsection{Analysis Emotion of Response.} To verify the effect of single and dual latent variables on the emotion of response, we present the generated responses with single or dual latent variables in Table~\ref{tab:table4}. In the first case, the introduction of single latent variable is less preferable due to the long context and fails to effectively capture the oneself emotion state (\emph{fantastic} and \emph{enjoying}) in the last turn. As a result, the response lacks self-emotion. In contrast, the introduction of dual latent variables allows for attending on both interlocutors' emotions and generating a more empathetic response.  
\vspace{0.0cm}
\begin{table}[t]
\centering
\begin{tabular}{lcccc}
\toprule
    \textbf{\ \ \ \ \ \ \ \ \ \ Models} & \textbf{Win} & \textbf{Loss} & \textbf{Tie} & \textbf{$\kappa$}\\
\midrule
    \textbf{CAB} vs MOEL & 35.7\% & 27.3\% & 40.0\% & 0.27 \\
    \textbf{CAB} vs MIME & 37.5\% & 25.0\% & 37.5\% & 0.35 \\
    \textbf{CAB} vs EmpDG & 38.3\% & 24.9\% & 36.7\% & 0.36 \\
    \textbf{CAB} vs KEMP & 43.5\% & 24.5\% & 32.0\% & 0.42 \\
    \textbf{CAB} vs CEM & 33.0\% & 31.0\% & 36.0\% & 0.47 \\
\bottomrule    
\end{tabular}
\caption{Results of Human A/B test, where 0.2\textless $\kappa$\textless 0.4 and 0.4\textless $\kappa$\textless 0.6 indicate fair and moderate agreements, respectively.}
\label{tab:table3}
\vspace{-0.5cm}
\end{table}

\begin{table}[t]
\vspace{0.4cm}
\centering
\begin{tabular}{|l|}
\hline
\textbf{Context}\\
{\color{Orange}{\textbf{Speaker}}}: I am looking forward to going on vacation in \\ a few weeks! We have a condo reserved on the beach, \\ with fantastic ocean views. I am ready!\ \ \ \ [\textbf{\emph{Anticipating}}]\\  {\color{RoyalBlue}{\textbf{Listener}}}: Ah, that sounds fantastic! Which ocean will \\ you be enjoying?\ \ \ \ \ \ \ \ \ \ \ \ \ \ \ \ \ \ \ \ \ \ \ \ \ \ \ \ \ \ \ \ \ \ \ \ \ \ \ \  [\textbf{\emph{Excited}}]\\{\color{Orange}{\textbf{Speaker}}}: Well, we are staying at panama city beach, so\\ we're right there at the gulf of mexico.\ \ \ \ \ \  [\textbf{\emph{Anticipating}}] \\
\hline
\textbf{Single Latent Variable}\\

\textbf{CAB}: I am sure you will be a great time!\\
\hline
\textbf{Dual Latent Variables}\\

\textbf{CAB}: That is awesome! I hope you have a great time to\\ have a vacation.\\
\hline
\end{tabular}
\caption{The responses generated with the single or dual latent variables. Bold words in brackets represent the interlocutors' emotions.}
\label{tab:table4}
\vspace{-0.3cm}
\end{table}

\subsection{Case Study}
The responses generated by CAB and other models are shown in Table~\ref{tab:table5}. In the first case, the response generated by MOEL is not fluent and leads to a worse user experience. KEMP and CEM only respond to the situation described by the user and neglect the user's feeling. Although both MIME and EMPDG acknowledge the user's feeling through 'That is awesome!', they only express empathy in affection and behavior. Through building paths between \textless \emph{kid}, \emph{excite}\textgreater\ and \textless \emph{disneyland}, \emph{time}\textgreater, CAB improves the understanding of the context and focuses on disneyland during decoding. The generated response can not only acknowledge the user's feeling but also be appropriate in emotion and content.

In other models, MOEL, EMPDG and CEM pay more attention to 'Everything turned out fine.' and ignore both interlocutors' emotions from the last turn. Therefore, they all generate positive replies. Although MIME may be proper in emotion, the response is generic and not semantically relevant. CAB, while recognizing the user's feeling, distinguishes the emotions of oneself and others, and generates 'That is a scary situation.' to acknowledge the user's situation and 'I hope you are ok' which may have a better comforting effect.

\begin{table}[t]


\centering
\begin{tabular}{c|l}
\toprule
\multirow{2}{*}{\textbf{Context}} & I was soooo excited to take our kids to \\ & disneyland for the first time! [\textbf{\emph{Excited}}]\\
\midrule
\multirow{2}{*}{Gold}  & Ooo they are gonna love it, just its expensive,\\ & I remember when it  was only \$20 to get in.\\
MOEL & That is great! I have never been to one of \\ & my favorite.\\
MIME & That is awesome! I bet you are excited!\\
EMPDG & That is awesome! I bet you are so excited!\\
KEMP & I bet that was a great time.\\
CEM & That sounds like a lot of fun!\\

\multirow{2}{*}{Know.} & kid, toy, amusement, \textbf{delight}, \textbf{thrill}, \textbf{excite}, \\ & \textbf{disneyland}, tourist, \textbf{enjoy}, time\\
Act & Acknowledging \\
\textbf{CAB} & That is \textbf{awesome}! I love \textbf{disneyland}!\\

\midrule
\multirow{8}{*}{\textbf{Context}} & Speaker: I recently had a scary experience. \\ & I heard gunshots on my street. [\textbf{\emph{Afraid}}]\\
 & Listener: Oh my! You must have been\\ & terrified! Did everything turn out ok? \\ &  [\textbf{\emph{Terrified}}]\\
& Speaker: Everything turned out fine. I think\\ & somebody in the nearby neighborhoods may \\ & have been hurt, though. [\textbf{\emph{Afraid}}]
\\
\midrule
\multirow{2}{*}{Gold}  & That is not good at all. That is a very\\ & scary situation.\\
MOEL & I am glad you are okay.\\

MIME & I am sorry to hear that. I hope you get it!\\
EMPDG & I am glad you are okay.\\
KEMP & I would be too. I would be too!\\
CEM & I am glad you are ok\\

\multirow{2}{*}{Know.} &  \textbf{street}, \textbf{city}, area,  \textbf{neighborhood}, nearby, \\ & around,  \textbf{turn}\\
Act  & Acknowledging \\
\textbf{CAB} & That is a \textbf{scary situation}. I hope you are ok!\\

\bottomrule
\end{tabular}
\caption{Responses generated by CAB and other models in different emotion states. Bold words in brackets represent the interlocutors' emotions and other bold words represent keywords and knowledge related to the dialogue context and responses.}
\label{tab:table5}
\end{table}
\section{Conclusions}
In this paper, we build paths by leveraging commonsense knowledge to enhance understanding of the user's situation, considering both interlocutors' emotions and guiding responses generation through dialogue act, namely by generating empathetic responses from three perspectives: cognition, affection and behavior. Extensive experiments based on both benchmark metrics and case studies have shown that our method CAB outperforms the state-of-the-art methods, demonstrating the effectiveness of our method in improving empathy of the generated responses.
\section{Acknowledgments}
This work was supported by the National Natural Science Foundation of China (61672144, 61872072).

\section{References}
\label{sec:reference_examples}

\nobibliography*
\bibentry{c:2016BVContinuous}.\\[.2em]
\bibentry{c:2022EmpHi}.\\[.2em]
\bibentry{c:2018DLGeneration}.\\[.2em]
\bibentry{hcr:2018EBTherapist}.\\[.2em] 
\bibentry{c:2021GLImCause}.\\[.2em]
\bibentry{c:2019GMEmotionRecognition}.\\[.2em]
\bibentry{hcr:1977GEmpathy}.\\[.2em]
\bibentry{hcr:2021KVEmoBERTa}.\\[.2em]
\bibentry{c:2015KBOptimization}.\\[.2em]
\bibentry{c:2014KWAuto-Encoding}.\\[.2em]
\bibentry{c:2016LGDiversity-Promoting}.\\[.2em]
\bibentry{c:2020LCEmpDG}.\\[.2em]
\bibentry{c:2022LLBridging}.\\[.2em]
\bibentry{c:2019LMMoEL}.\\[.2em]
\bibentry{hcr:2020LWDiverse}.\\[.2em]
\bibentry{hcr:2009LWreview}.\\[.2em]
\bibentry{c:2021LZSupport}.\\[.2em]
\bibentry{hcr:2021LMGPT}.\\[.2em]
\bibentry{c:2021LDResponses}.\\[.2em]
\bibentry{c:2020MHMIME}.\\[.2em]
\bibentry{c:2004MTTextRank}.\\[.2em]
\bibentry{hcr:2007MDEmpathy}.\\[.2em]
\bibentry{c:2014PSGlove}.\\[.2em]
\bibentry{c:2019RSBenchmark}.\\[.2em]
\bibentry{hcr:2021RLAutoencoder}.\\[.2em]
\bibentry{c:2022SZCEM}.\\[.2em]
\bibentry{c:2015SLCVAE}.\\[.2em]
\bibentry{c:2017VSAttention}.\\[.2em]
\bibentry{c:2015VFNetworks}.\\[.2em]
\bibentry{hcr:2015VLConversational}.\\[.2em]
\bibentry{hcr:2021WLMultiCausality}.\\[.2em]
\bibentry{hcr:2021WWCASS}.\\[.2em]
\bibentry{c:2020WPTaxonomy}.\\[.2em]
\bibentry{c:2020WLContext-Specific}.\\[.2em]
\bibentry{c:2017ZZDiscourse-level}.\\[.2em]
\bibentry{c:2021ZLCOMAE}.\\[.2em]

\nobibliography{aaai22}


\newpage

\appendix
\begin{appendices}
\subsection{A Data Annotation}
\textbf{Emotion} is annotated by a classfier of EmoBERTa \cite{hcr:2021KVEmoBERTa} which is fine-tuned with situation sentences. The classifier achieved a micro F1 of 63.14\% on the \emph{EmpatheticDialogues} test dataset. We applied the obtained classifier on all listener utterances. The predicted emotion label is one of the 32 categories.

\textbf{Dialogue Act} is annotated by EmoBERT which is prompted by Welivita and Pu (\citeyear{c:2020WPTaxonomy}). We first finetuned EmoBERT on a subset of \emph{EmpatheticDialogues} dataset, which contains dialogue act of 8 categories (\emph{agreeing}, \emph{acknowledging}, \emph{encouraging}, \emph{consoling}, \emph{sympathizing}, \emph{suggesting}, \emph{questioning}, \emph{wishing}). Then the fine-tuned EmoBERT was used to label all listener's utterances and the final classification accuracy reached 94.15\%, indicating that the annotation model has a reliable performance.
\subsection{B Knowledge Acquisition and Fusion}
We first use the TextRank algorithm \cite{c:2004MTTextRank} to extract 3-10 important keywords based on the length of the speaker's utterances, and then filter words with noun, verb, adverb and adjective part of speech among these keywords to keep $\textbf{cw}$ keywords. Next, we construct the removed relation set (\emph{ExternalURL}, \emph{NotDesires}, \emph{NotHasProperty}, \emph{NotCapableOf}, \emph{dbpedia}, \emph{DistinctFrom}, \emph{EtymologicallyDerivedFrom}, \emph{EtymologicallyRelatedTo}, \emph{SymbolOf}, \emph{FormOf}, \emph{AtLocation}, \emph{DerivedFrom}, \emph{CreatedBy}, \emph{MadeOf}) based on Li et al. (\citeyear{c:2022LLBridging}) in which the relations are removed among 38 relations, and the mapping of relations to natural language is shown in Table~\ref{tab:table7}, where 'Reverse-' denotes the relationship after reversing the head and tail entities. 

The dialogues in which no path is found have a special process. Similarly, we first change the values of $K, k, Num$ and then repeat Step a and b (please refer to the Knowledge Acquisition subsection for Step a and b). Then, the remaining 4.8\% of the data in which the path is still not found is supplemented by finding a one-hop knowledge triple according to Step a.
The relevant hyper-parameter are set as in Table~\ref{tab:table6}.

\begin{table}[H]

\centering
\begin{tabular}{ccc}
\toprule
\textbf{Hyper-parameters} & 
\textbf{First} & \textbf{Second} \\
\midrule
$M$ & 4 & 1\\
$K$ & 5 & 1\\
$k$ & 3 & 1 \\
$Num$ & 15 & 15\\
\bottomrule
\end{tabular}
\caption{Hyper-parameters setting for searching the paths}
\label{tab:table6}
\end{table}

\begin{table*}[t]
\centering

\begin{tabular}{cccc}
\toprule
\textbf{Relationship} & 
\textbf{Natural Language} & \textbf{Relationship} & \textbf{Natural Language}\\
\midrule
IsA & {is a} & {Reverse-IsA} & {is a}\\
HasProperty & can & Reverse-HasProperty & is an attribute of \\
Desires & desires & Reverse-Desires & is desired by\\
HasA & has & Reverse-HasA & is owned by\\
RelatedTo & is related to & Reverse-RelatedTo & is related to\\
ReceivesAction & can be & Reverse-ReceivesAction & is\\
Causes & causes & Reverse-Causes & is because of\\
HasSubevent & then & Reverse-HasSubevent & before\\
UsedFor & is used for & Reverse-UsedFor & needs\\
PartOf & is part of & Reverse-PartOf & includes\\
HasPrerequisite & has prerequisite & Reverse-HasPrerequisite & is the condition of\\
HasContext & has meaning of & Reverse-HasContext & has meaning of\\
MannerOf & is one manner of & Reverse-MannerOf & is the result of\\
SimilarTo & is similar to & Reverse-SimilarTo & is similar tois similar to\\
CapableOf & can & Reverse-CapableOf & benefit from\\
MotivatedByGoal & becauses & Reverse-MotivatedByGoal & desires\\
CausesDesire & desires & Reverse-CausesDesire & is desired by\\
LocatedNear & is located near & Reverse-LocatedNear & is located near\\
Entails & entails & Reverse-Entails & is part of\\
HasLastSubevent & then & Reverse-HasLastSubevent & before\\
HasFirstSubevent & then & Reverse-HasFirstSubevent & before\\
Antonym & is opposite to & Reverse-Antonym & is opposite to\\
Synonym & is similar to & Reverse-Synonym & is similar to\\
\bottomrule
\end{tabular}
\caption{Rules of relation transformation}
\label{tab:table7}
\end{table*}

\end{appendices}


\begin{thebibliography}{37}
\providecommand{\natexlab}[1]{#1}

\bibitem[{Bowman et~al.(2016)Bowman, Vilnis, Vinyals, Dai, J{\'{o}}zefowicz,
  and Bengio}]{c:2016BVContinuous}
Bowman, S.~R.; Vilnis, L.; Vinyals, O.; Dai, A.~M.; J{\'{o}}zefowicz, R.; and
  Bengio, S. 2016.
\newblock {Generating Sentences from a Continuous Space}.
\newblock In \emph{Proceedings of the 20th {SIGNLL} Conference on Computational
  Natural Language Learning}, 10--21.

\bibitem[{Chen, Li, and Yang(2022)}]{c:2022EmpHi}
Chen, M.~Y.; Li, S.; and Yang, Y. 2022.
\newblock {EmpHi: Generating Empathetic Responses with Human-like Intents}.
\newblock In \emph{Proceedings of the 2022 Conference of the North American
  Chapter of the Association for Computational Linguistics: Human Language
  Technologies}, 1063--1074.

\bibitem[{Du et~al.(2018)Du, Li, He, Xu, Bing, and Wang}]{c:2018DLGeneration}
Du, J.; Li, W.; He, Y.; Xu, R.; Bing, L.; and Wang, X. 2018.
\newblock {Variational Autoregressive Decoder for Neural Response Generation}.
\newblock In \emph{Proceedings of the 2018 Conference on Empirical Methods in
  Natural Language Processing}, 3154--3163.

\bibitem[{Elliott et~al.(2018)Elliott, Bohart, Watson, and
  Murphy}]{hcr:2018EBTherapist}
Elliott, R.; Bohart, A.~C.; Watson, J.~C.; and Murphy, D. 2018.
\newblock Therapist empathy and client outcome: An updated meta-analysis.
\newblock \emph{Psychotherapy}, 55(4): 399–410.

\bibitem[{Gao et~al.(2021)Gao, Liu, Deng, Wang, Cao, Du, and
  Xu}]{c:2021GLImCause}
Gao, J.; Liu, Y.; Deng, H.; Wang, W.; Cao, Y.; Du, J.; and Xu, R. 2021.
\newblock {Improving Empathetic Response Generation by Recognizing Emotion
  Cause in Conversations}.
\newblock In \emph{Findings of the Association for Computational Linguistics:
  {EMNLP} 2021}, 807--819.

\bibitem[{Ghosal et~al.(2019)Ghosal, Majumder, Poria, Chhaya, and
  Gelbukh}]{c:2019GMEmotionRecognition}
Ghosal, D.; Majumder, N.; Poria, S.; Chhaya, N.; and Gelbukh, A.~F. 2019.
\newblock {DialogueGCN: {A} Graph Convolutional Neural Network for Emotion
  Recognition}.
\newblock In \emph{Proceedings of the 2019 Conference on Empirical Methods in
  Natural Language Processing and the 9th International Joint Conference on
  Natural Language Processing}, 154--164.

\bibitem[{Gladstein(1977)}]{hcr:1977GEmpathy}
Gladstein, G.~A. 1977.
\newblock Empathy and Counseling Outcome: An Empirical and Conceptual Review.
\newblock \emph{Counseling Psychologist}, 6(4): 70--79.

\bibitem[{Kim and Vossen(2021)}]{hcr:2021KVEmoBERTa}
Kim, T.; and Vossen, P. 2021.
\newblock EmoBERTa: Speaker-Aware Emotion Recognition in Conversation with
  RoBERTa.
\newblock \emph{CoRR}, abs/2108.12009.

\bibitem[{Kingma and Ba(2019)}]{c:2015KBOptimization}
Kingma, D.~P.; and Ba, J. 2019.
\newblock {Adam: {A} Method for Stochastic Optimization}.
\newblock In \emph{3rd International Conference on Learning Representations}.

\bibitem[{Kingma and Welling(2014)}]{c:2014KWAuto-Encoding}
Kingma, D.~P.; and Welling, M. 2014.
\newblock {Auto-Encoding Variational Bayes}.
\newblock In \emph{2nd International Conference on Learning Representations}.

\bibitem[{Li et~al.(2016)Li, Galley, Brockett, Gao, and
  Dolan}]{c:2016LGDiversity-Promoting}
Li, J.; Galley, M.; Brockett, C.; Gao, J.; and Dolan, B. 2016.
\newblock {A Diversity-Promoting Objective Function for Neural Conversation
  Models}.
\newblock In \emph{The 2016 Conference of the North American Chapter of the
  Association for Computational Linguistics: Human Language Technologies},
  110--119.

\bibitem[{Li et~al.(2020)Li, Chen, Ren, Ren, Tu, and Chen}]{c:2020LCEmpDG}
Li, Q.; Chen, H.; Ren, Z.; Ren, P.; Tu, Z.; and Chen, Z. 2020.
\newblock {EmpDG: Multi-resolution Interactive Empathetic Dialogue Generation}.
\newblock In \emph{Proceedings of the 28th International Conference on
  Computational Linguistics}, 4454--4466.

\bibitem[{Li et~al.(2022)Li, Li, Ren, Ren, and Chen}]{c:2022LLBridging}
Li, Q.; Li, P.; Ren, Z.; Ren, P.; and Chen, Z. 2022.
\newblock {Knowledge Bridging for Empathetic Dialogue Generation}.
\newblock In \emph{Thirty-Sixth {AAAI} Conference on Artificial Intelligence},
  10993--11001.

\bibitem[{Lin et~al.(2021)Lin, Madotto, Shin, Xu, and Fung}]{c:2019LMMoEL}
Lin, Z.; Madotto, A.; Shin, J.; Xu, P.; and Fung, P. 2021.
\newblock {MoEL: Mixture of Empathetic Listeners}.
\newblock In \emph{Proceedings of the 2019 Conference on Empirical Methods in
  Natural Language Processing and the 9th International Joint Conference on
  Natural Language Processing}, 121--132.

\bibitem[{Lin et~al.(2020)Lin, Winata, Xu, Liu, and Fung}]{hcr:2020LWDiverse}
Lin, Z.; Winata, G.~I.; Xu, P.; Liu, Z.; and Fung, P. 2020.
\newblock Variational Transformers for Diverse Response Generation.
\newblock \emph{CoRR}, abs/2003.12738.

\bibitem[{Liu et~al.(2009)Liu, Wang, Yu, and Wang}]{hcr:2009LWreview}
Liu, C.; Wang, Y.; Yu, G.; and Wang, Y. 2009.
\newblock A review of relevant theories of empathy and exploration of new
  dynamic models.
\newblock \emph{Advances in Psychological Science}, (5): 9.

\bibitem[{Liu et~al.(2021{\natexlab{a}})Liu, Zheng, Demasi, Sabour, and
  Huang}]{c:2021LZSupport}
Liu, S.; Zheng, C.; Demasi, O.; Sabour, S.; and Huang, M. 2021{\natexlab{a}}.
\newblock {Towards Emotional Support Dialog Systems}.
\newblock In \emph{Proceedings of the 59th Annual Meeting of the Association
  for Computational Linguistics and the 11th International Joint Conference on
  Natural Language Processing}, 3469--3483.

\bibitem[{Liu et~al.(2021{\natexlab{b}})Liu, Du, Li, and
  Xu}]{c:2021LDResponses}
Liu, Y.; Du, J.; Li, X.; and Xu, R. 2021{\natexlab{b}}.
\newblock {Generating Empathetic Responses by Injecting Anticipated Emotion}.
\newblock In \emph{{IEEE} International Conference on Acoustics, Speech and
  Signal Processing}, 7403--7407.

\bibitem[{Liu et~al.(2021{\natexlab{c}})Liu, Maier, Minker, and
  Ultes}]{hcr:2021LMGPT}
Liu, Y.; Maier, W.; Minker, W.; and Ultes, S. 2021{\natexlab{c}}.
\newblock Empathetic Dialogue Generation with Pre-trained RoBERTa-GPT2 and
  External Knowledge.
\newblock \emph{CoRR}, abs/2109.03004.

\bibitem[{Majumder et~al.(2020)Majumder, Hong, Peng, Lu, and
  Poria}]{c:2020MHMIME}
Majumder, N.; Hong, P.; Peng, S.; Lu, J.; and Poria, S. 2020.
\newblock {MIME: MIMicking Emotions for Empathetic Response Generation}.
\newblock In \emph{Proceedings of the 2020 Conference on Empirical Methods in
  Natural Language Processing}, 8968--8979.

\bibitem[{Mihalcea and Tarau(2004)}]{c:2004MTTextRank}
Mihalcea, R.; and Tarau, P. 2004.
\newblock {TextRank: Bringing Order into Text}.
\newblock In \emph{Proceedings of the 2004 Conference on Empirical Methods in
  Natural Language Processing}, 404--411.

\bibitem[{Moriguchi et~al.(2007)Moriguchi, Decety, Ohnishi, Maeda, Mori,
  Nemoto, Matsuda, and Komaki}]{hcr:2007MDEmpathy}
Moriguchi, Y.; Decety, J.; Ohnishi, T.; Maeda, M.; Mori, T.; Nemoto, K.;
  Matsuda, H.; and Komaki, G. 2007.
\newblock Empathy and judging other's pain: an fMRI study of alexithymia.
\newblock \emph{Cerebral Cortex}, 17(9): 2223--2234.

\bibitem[{Paszke et~al.(2019)Paszke, Gross, Massa, Lerer, Bradbury, Chanan,
  Killeen, Lin, Gimelshein, Antiga, Desmaison, K{\"{o}}pf, Yang, DeVito,
  Raison, Tejani, Chilamkurthy, Steiner, Fang, Bai, and
  Chintala}]{c:2019PGPyTorch}
Paszke, A.; Gross, S.; Massa, F.; Lerer, A.; Bradbury, J.; Chanan, G.; Killeen,
  T.; Lin, Z.; Gimelshein, N.; Antiga, L.; Desmaison, A.; K{\"{o}}pf, A.; Yang,
  E.~Z.; DeVito, Z.; Raison, M.; Tejani, A.; Chilamkurthy, S.; Steiner, B.;
  Fang, L.; Bai, J.; and Chintala, S. 2019.
\newblock {PyTorch: An Imperative Style, High-Performance Deep Learning
  Library}.
\newblock In \emph{Advances in Neural Information Processing Systems 32: Annual
  Conference on Neural Information Processing Systems 2019}, 8024--8035.

\bibitem[{Pennington, Socher, and Manning(2014)}]{c:2014PSGlove}
Pennington, J.; Socher, R.; and Manning, C.~D. 2014.
\newblock {Glove: Global Vectors for Word Representation}.
\newblock In \emph{Proceedings of the 2014 Conference on Empirical Methods in
  Natural Language Processing}, 1532--1543.

\bibitem[{Rashkin et~al.(2019)Rashkin, Smith, Li, and
  Boureau}]{c:2019RSBenchmark}
Rashkin, H.; Smith, E.~M.; Li, M.; and Boureau, Y. 2019.
\newblock {Towards Empathetic Open-domain Conversation Models: {A} New
  Benchmark and Dataset}.
\newblock In \emph{Proceedings of the 57th Conference of the Association for
  Computational Linguistics}, 5370--5381.

\bibitem[{Ruan and Ling(2021)}]{hcr:2021RLAutoencoder}
Ruan, Y.; and Ling, Z. 2021.
\newblock Emotion-Regularized Conditional Variational Autoencoder for Emotional
  Response Generation.
\newblock \emph{CoRR}, abs/2104.08857.

\bibitem[{Sabour, Zheng, and Huang(2022)}]{c:2022SZCEM}
Sabour, S.; Zheng, C.; and Huang, M. 2022.
\newblock {{CEM:} Commonsense-Aware Empathetic Response Generation}.
\newblock In \emph{Thirty-Sixth {AAAI} Conference on Artificial Intelligence},
  11229--11237.

\bibitem[{Sohn, Lee, and Yan(2015)}]{c:2015SLCVAE}
Sohn, K.; Lee, H.; and Yan, X. 2015.
\newblock {Learning Structured Output Representation using Deep Conditional
  Generative Models}.
\newblock In \emph{Advances in Neural Information Processing Systems 28: Annual
  Conference on Neural Information Processing Systems 2015}, 3483--3491.

\bibitem[{Vaswani et~al.(2017)Vaswani, Shazeer, Parmar, Uszkoreit, Jones,
  Gomez, Kaiser, and Polosukhin}]{c:2017VSAttention}
Vaswani, A.; Shazeer, N.; Parmar, N.; Uszkoreit, J.; Jones, L.; Gomez, A.~N.;
  Kaiser, L.; and Polosukhin, I. 2017.
\newblock {Attention is All you Need}.
\newblock In \emph{Advances in Neural Information Processing Systems 30: Annual
  Conference on Neural Information Processing Systems 2017}, 5998--6008.

\bibitem[{Vinyals, Fortunato, and Jaitly(2015)}]{c:2015VFNetworks}
Vinyals, O.; Fortunato, M.; and Jaitly, N. 2015.
\newblock {Pointer Networks}.
\newblock In \emph{Advances in Neural Information Processing Systems 28: Annual
  Conference on Neural Information Processing Systems 2015}, 2692--2700.

\bibitem[{Vinyals and Le(2015)}]{hcr:2015VLConversational}
Vinyals, O.; and Le, Q.~V. 2015.
\newblock A Neural Conversational Model.
\newblock \emph{CoRR}, abs/1506.05869.

\bibitem[{Wang et~al.(2021{\natexlab{a}})Wang, Li, Lin, and
  Mu}]{hcr:2021WLMultiCausality}
Wang, J.; Li, W.; Lin, P.; and Mu, F. 2021{\natexlab{a}}.
\newblock Empathetic Response Generation through Graph-based Multi-hop
  Reasoning on Emotional Causality.
\newblock \emph{Knowl. Based Syst.}, 233: 107547.

\bibitem[{Wang et~al.(2021{\natexlab{b}})Wang, Wang, Tian, Peng, Fan, Zhang,
  Yu, Ma, and Wang}]{hcr:2021WWCASS}
Wang, L.; Wang, D.; Tian, F.; Peng, Z.; Fan, X.; Zhang, Z.; Yu, M.; Ma, X.; and
  Wang, H. 2021{\natexlab{b}}.
\newblock CASS: Towards Building a Social-Support Chatbot for Online Health
  Community.
\newblock \emph{Proc. {ACM} Hum. Comput. Interact.}, 5({CSCW1}): 1--31.

\bibitem[{Welivita and Pu(2020)}]{c:2020WPTaxonomy}
Welivita, A.; and Pu, P. 2020.
\newblock {A Taxonomy of Empathetic Response Intents in Human Social
  Conversations}.
\newblock In \emph{Proceedings of the 28th International Conference on
  Computational Linguistics}, 4886--4899.

\bibitem[{Wu et~al.(2020)Wu, Li, Zhang, Zhou, and
  Wu}]{c:2020WLContext-Specific}
Wu, S.; Li, Y.; Zhang, D.; Zhou, Y.; and Wu, Z. 2020.
\newblock {Diverse and Informative Dialogue Generation with Context-Specific
  Commonsense Knowledge Awareness}.
\newblock In \emph{Proceedings of the 58th Annual Meeting of the Association
  for Computational Linguistics}, 5811--5820.

\bibitem[{Zhao, Zhao, and Esk{\'{e}}nazi(2017)}]{c:2017ZZDiscourse-level}
Zhao, T.; Zhao, R.; and Esk{\'{e}}nazi, M. 2017.
\newblock {Learning Discourse-level Diversity for Neural Dialog Models using
  Conditional Variational Autoencoders}.
\newblock In \emph{Proceedings of the 55th Annual Meeting of the Association
  for Computational Linguistics}, 654--664.

\bibitem[{Zheng et~al.(2021)Zheng, Liu, Chen, Leng, and Huang}]{c:2021ZLCOMAE}
Zheng, C.; Liu, Y.; Chen, W.; Leng, Y.; and Huang, M. 2021.
\newblock {CoMAE: {A} Multi-factor Hierarchical Framework for Empathetic
  Response Generation}.
\newblock In \emph{Findings of the Association for Computational Linguistics:
  {ACL/IJCNLP}}, 813--824.

\end{thebibliography}
\end{document}